\documentclass[11pt]{article}
\usepackage{coling2020}
\usepackage{times}
\usepackage{url}
\usepackage{latexsym}
\usepackage{graphicx}

\colingfinalcopy 

\title{IIT\_kgp at FinCausal 2020, Shared Task 1: Causality Detection using Sentence Embeddings in Financial Reports}

\author{Arka Mitra\\
Electronics and \\
Electrical Communication \\ Engineering\\
  IIT Kharagpur \\
  {\tt thearkamitra}\\
  {\tt @iitkgp.ac.in} \\\And
  Harshvardhan Srivastava \\
  Electrical   Engineering \\
  IIT Kharagpur \\
  {\tt hvs24}\\
  {\tt @iitkgp.ac.in} \\\And
  Yugam Tiwari\\
  Mechanical Engineering\\
  IIT Kharagpur \\
  {\tt tiwariyugam}\\
  {\tt @iitkgp.ac.in}\\
  }
\date{}

\begin{document}

\maketitle
\begin{abstract}
  The paper describes the work that the team submitted to FinCausal 2020 Shared Task. This work is associated with the first sub-task of identifying causality in sentences. The various models used in the experiments tried to obtain a latent space representation for each of the sentences. Linear regression was performed on these representations to classify whether the sentence is causal or not. The experiments have shown BERT (Large)  performed the best, giving a F1 score of 0.958, in the task of detecting the causality of sentences in financial texts and reports. The class imbalance was dealt with a modified loss function to give a better metric score for the evaluation.
\end{abstract}

\section{Introduction}
\label{intro}

%
%
\blfootnote{
    %
    %
    \hspace{-0.65cm}  
    %
    
    
    
    \hspace{-0.65cm}  
    This work is licensed under a Creative Commons 
    Attribution 4.0 International License.
    License details:
    \url{http://creativecommons.org/licenses/by/4.0/}.
}

The last few decades have seen an advent in the storage capacities which has inherently increased the amount of data that people can store. This generates a massive amount of data as well. One such type of data which is of utmost importance is financial data comprising mainly of financial reports. It is not possible to go through all the reports as the number of such reports keep on increasing with time and it is expensive to hire people to summarize the main contents. But these data can contain important information which might help in explaining the profits gained or losses incurred by a branch or a company as a whole. It also provides key insights to the company's decisions and it's monetary effect on the stock prices of the company. Thus there is also a need for more and more financial analysis. In order to explain the variability in the data and to draw conclusions from it, we need to identify the causal relationships between various sentences occurring in the document.\\
A causal relationship implies that there is an underlying dependency between the two clauses of the sentence. If \textit{cause} clause triggers the \textit{effect} clause in the sentence and the effect can be explained by the cause, this implies that sentence is causal. The trigger word that can be used to determine and separate the two clauses can be implicit like verbs, propositions, conjunctions etc. or can be explicit that can be understood from the overall structure of the sentence. But the triggers can be present in multiple positions and all of them might not contribute to the main causal relationship present in the sentence. Also, the presence of these triggers does not always imply that the sentence has causality. Thus it is important to improve from regular expression based methods and introduce a new method.\\
Causality has already been studied extensively in a general field but it has not been extended to the field of financial data considerably. The FinCausal 2020 tries to extend the same methodologies in the financial domain and see their performance to have a more comprehensive understanding of what works better in the domain of finance. The challenges posed by financial narratives are at the heart of a lot of discussions within the finance industry. As a Natural Language Generation provider with a specialized focus on Financial Reporting, Yseop  \cite{newpap} is often faced with the need to deliver a relevant mapping of events, indicators, and facts, making causality one of their main research topics.
The task we are tackling here can be considered analogous to a sequence classification task where we are required to assign labels to a sentence. This motivated us to use a transformer-based architecture which has shown state of the art results in many natural language tasks like text classification, language modelling, neural machine translation, etc.
In this Shared Task, the phenomenon of lexical causative \cite{LEVIN199435} is not taken into
account. A lexical causative is a causal relationship stated through specific connectives (generally predicates) which, from a semantic point of view, also bear the effect of the cause. We will not consider those as causal references, since the effects are implied in the connectives’ definition. For instance in ”The
company decreased its provisions in 2018.”, decrease is a lexical causative that can be glossed as make
something lower.\\
There can also be different types of relationships along with the causal relationship in the identified
text section. It is often rendered with the use of polysemous connectives whose main function is not
to introduce a causal relationship. For example, in this sentence: ”Zhao found himself 60 million yuan
indebted after losing 9,000 BTC in a single day (February 10, 2014)”, the main function of the connective
after is to express a temporal relation between the two clauses. But we also have a causal relationship
between them, since one triggers the other. \\
We are using the term text section since it could be a phrase, a sentence or a paragraph in which
the cause and the effect are split in different sentences.\\

\section{Data}
    The data was provided as a part of the Financial Narrative Workshop by the organisers \cite{Mariko-fincausal-2020}. The data was mainly provided as csv files.  
    The training data consisted of about 4200 sentences. The data was divided into two classes depending on whether they had causality in their underlying structure (causal sentence) or not (non-causal sentence). The data mainly consisted of non-causal sentences and the number of non-causal sentences was about 16 times that of causal sentences.
\section{Method}
\begin{figure}[h!]
    
\begin{minipage}[b]{.48\linewidth}

  \centering
  \centerline{\includegraphics[width=8.0cm]{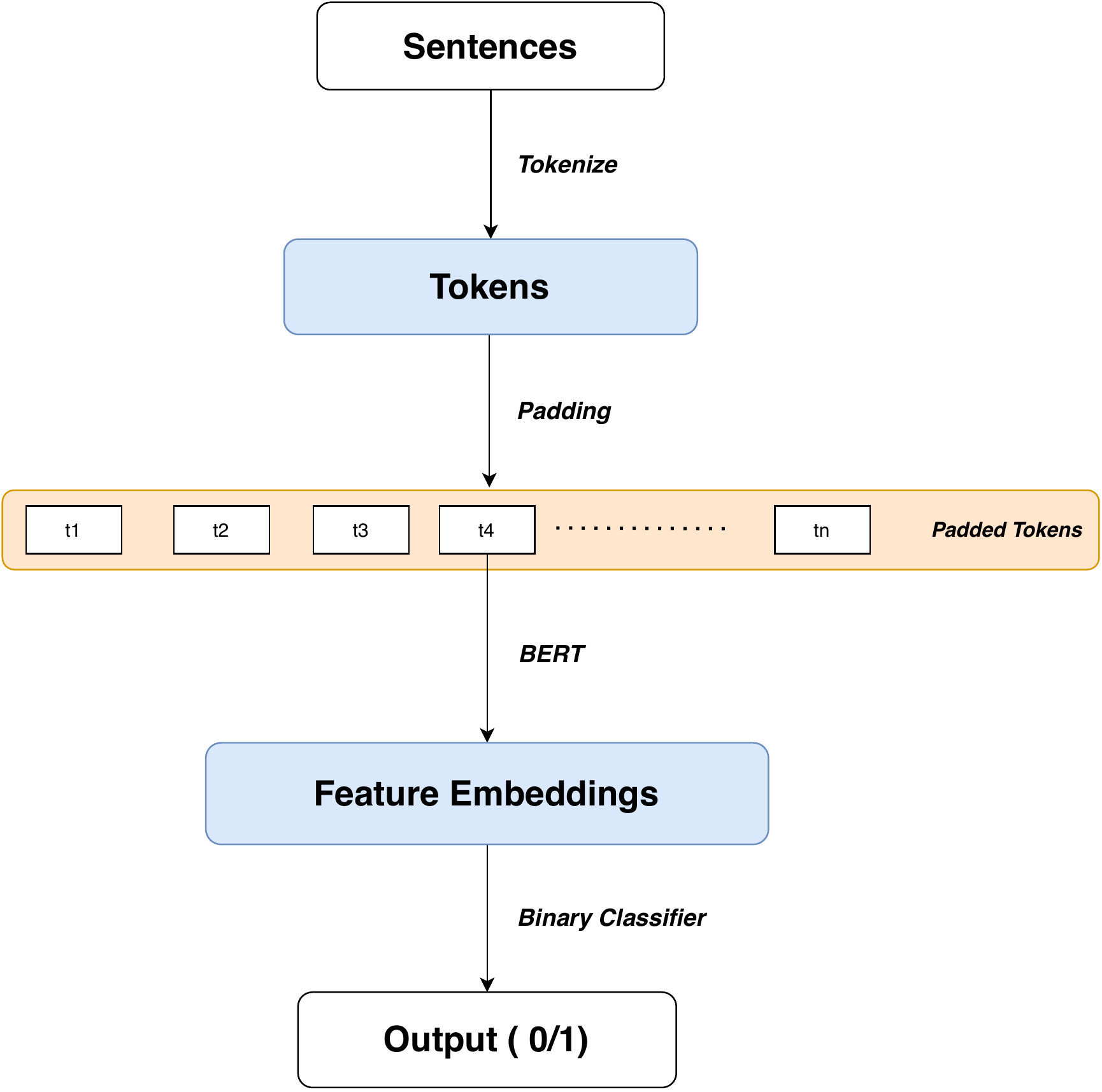}}
  \centerline{(a) Overall Pipeline}\medskip
\end{minipage}
\hfill
\begin{minipage}[b]{0.48\linewidth}
  \centering 
  \centerline{\includegraphics[width=4.0cm]{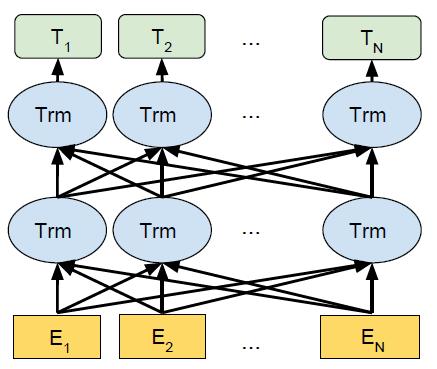}}
  \centerline{(b) BERT MODEL}\medskip
\end{minipage}
\caption{(a) shows the overall methodology that has been followed. (b) shows the architecture of the BERT Model that takes in a sentence and output the corresponding embeddings.}
\label{fig.1}
\end{figure}

\subsection{Motivation}
Earlier trigger words were identified for classifying sentences into causal and non-causal. However if we consider the word "Since" and the following examples, it will become quite clear why it is not always a correct way to do so.\\
    • "Since the deer could not run slow, the lion was able to catch it."\\
    • "Since morning, the man was not feeling well." \\
In the first case, the cause-effect relationship is quite visible. However, in the second case, a human can easily identify that the sentence does not have a causal relationship. One of the main reasons due to which humans are capable at doing so is because they can identify the context in which the words are spoken. With the advances in natural language processing, we have models which can also consider contextual representation.\\
This motivated us to try models like BERT \cite{devlin2018bert}, XLNet \cite{yang2019xlnet} and RoBERTa \cite{liu2019roberta}. They take into consideration the contextual embedding of each word. We generated the sentence embeddings of each sentence and then passed it through a linear layer to classify whether the sentence is causal or not. \ref{fig.1} shows the overall pipeline of our model. \\

\subsection{Loss}
The evaluation metric was taken to be the \textbf{F1} score which is the harmonic mean of \textit{precision} and \textit{recall}. When there is a high class imbalance in the dataset, the model generally has a high recall or precision. As stated in the data section, the number of non-causal sentences exceeded the number of causal sentences by an order magnitude and thus it has high precision but low recall. To tackle that, whenever there was a false negative the loss function was configured to penalize it heavily \cite{Ho_2020}. The loss \textit{L} is defined by the following:
\begin{equation}
    L = -\alpha*y_{true}*\log(y_{pred})  -(1-y_{true})*\log(1-y_{pred})
\end{equation}
The parameter alpha is introduced as a coefficient of the first term in the product. Increasing the value of alpha makes the model adjust it's weights thereby decreasing the number of false negatives. Hence the recall of the system increases but the precision decreases by only a small amount. This increases the overall F1 score.
\subsection{Backbone}
As a state of the art language model BERT (Bidirectional Encoder Representations from Transformers) has achieved great results in the task of text classification. Models like BERT takes in an input of atmost 512 sequential tokens and generates a vector representation of the sequence. We followed a similar methodology as BERT for text classification \cite{sun2019finetune}. After obtaining the BERT representation of the sentence, we added a linear classifier on top of it with an output of two nodes. A softmax was applied on the logits and the one with the max score was selected as the predicted output.
\begin{equation}
    y_{pred} = argmax(softmax(W*BERT(sentence) + b))
\end{equation}
where W is the Linear layer matrix and b is the corresponding bias.\\
We investigated different masked attention models like the BERT-uncased-base, BERT-cased-base, BERT-cased-large, XLNet-base, RoBERTa-cased-base for this sub-task and it was seen that the best results were obtained for the BERT-cased-large model. Cased models performed better than the uncased ones when tried out with the BERT base and thus in the consequent experiments, we proceeded with the cased versions. Other transformer-based language models like XLNet (large) and RoBERTa (large) were not employed due to their large size and our limited access to computational power.
\subsection{Input Tokens Length}
Another method that we tried to explore was the relationship between the input dimensions and the overall performance. \cite{tan2019efficientnet} shows that there exists a correlation between the width of the architecture, height of the architecture and the initial input shape which is fed into the network. Also, the paper had shown that when only one of the individual parameters was changed, the performance saturated very quickly. We had also changed the input dimension and the model height (by using BERT base and large). There was a trade-off between the input shape and the performance. Due to the large hidden dimension size, the Floating Point Operations (FLOPs) of the model is almost linearly proportional to the input length to the transformer \cite{dai2020funneltransformer}. Decrease in the number of FLOPs also decreases the time for completing the same number of epochs. However, if the input size is decreased, some important information may get lost due to the truncation which decreases the performance of the model. By some analysis of the data that had been provided, we saw that most sentences that were causal were at most 128 words long, and thereby it was chosen as the truncation length for the BERT model.
\begin{table}[h!]
    \centering
    \caption{Sequence length variation of BERT-base in experiments}
    \begin{tabular}{ |p{3cm}|p{3cm}|p{3cm}|p{3cm}|  }
    
     \hline
     \textbf{Sequence length}& \textbf{F1 Score} & \textbf{Precision}& \textbf{Recall}\\
     \hline
     64&0.94268&0.942527&0.942835\\
     128&0.948066&0.948282&0.947856\\
     256&0.951081&0.951650&0.950560\\
     512&0.951879&0.953650&0.950717\\
     \hline
    \end{tabular}
\label{Seq_length}
\end{table}

\section{Results}

The datasets that had been provided as a part of the competition had been given as Trial1, Practice1, and Evaluation1. To choose the model on which to proceed further, the initial experiments were run on Trail1 dataset. The F1 scores of the different models followed a ranking of BERT Large, XLNET-Base, BERT-Base, and RoBERTa with BERT-Large being the largest and the RoBERTa results being the lowest. The experiments had been run on Google Colaboratory \textbf{using a single 12 GB NVIDIA K80 Tesla GPU} and due to computing limitations, XLNet Large and RoBERTa Large could not be calculated.
\begin{table}[h!]
    \centering
    \caption{Results of the Different Experiments}
    \begin{tabular}{ |p{3cm}|p{3cm}|p{3cm}|p{3cm}|  }
    
     \hline
     \textbf{Model}& \textbf{F1 Score} & \textbf{Precision }& \textbf{Recall}\\
     \hline
     XLNet (Base)&0.950820&0.952041&0.949788\\
     
     RoBERTa&0.929485&0.871544&0.921205\\
     
     BERT (Base) &0.948066&0.948282&0.947856 \\
     
     BERT (Large)&0.957814&0.957408&0.958299\\
     \hline
    \end{tabular}
\label{tab:my_label}
\end{table}
Based on these results, the outputs from BERT-Large-Cased had been sent into the final evaluation system which resulted in an \textbf{F1} score of \textbf{0.958} resulting in us securing the 8$^{th}$ position.
\section{Discussion}
This method uses contextual word embeddings and removes most of the problems a regular expression based method would have had.
Based on the results, a comparative study was done on the performance of various models like BERT, XLNet and RoBERTa. The BERT-large model performs better than BERT-base as expected as it can have more hidden features to explore. XLNet performs better than BERT as the former is a mix of both masked and permutation language modeling. RoBERTa performed worse than BERT despite it being trained on more data. One of the reasons for that might be as RoBERTa is not trained on next sentence prediction (NSP). Better results are expected on cased versions of language models as compared to the uncased versions. Uncased versions turn all the sentences to lowercase which might lead to loss of information in some cases. The added weights on the loss function also helped to increase the recall and thereby helped to tackle the class imbalance.\\

\section{Conclusion}
We had experimented with various models and compared the relative accuracy on the given dataset.
We finally submitted the best results on a given part of the dataset due to computation limitations and obtained an F1 score 0f 0.958. We can conclude that the cased models performed better than uncased models in all scenarios that we encountered. 
\section{Future Works}
The BERT Large Cased model had given relatively good results. However, a domain specific BERT would have provided better results \cite{gururangan2020dont}. Also, we had used just a part of the whole dataset due to restricted access to computational power. Including a larger dataset would also help to even improve the scores.
\bibliographystyle{coling}
\bibliography{coling2020}

\end{document}